\documentclass{scrartcl}
\usepackage{tikz}
\usetikzlibrary{positioning,arrows.meta,shapes}
\usepackage{subcaption}
\usepackage{booktabs}

\usepackage{geometry}

\usepackage{blindtext}
\usepackage[T1]{fontenc}
\usepackage[utf8]{inputenc}

\usepackage{amsmath, amsfonts, amsthm, amssymb}
\usepackage{braket, nicefrac}

\usepackage{siunitx}

\usepackage{enumitem, multicol}

\usepackage{graphicx, float}  % Use option [H] to force the placement of a figure
\usepackage{keystroke}
\usepackage{pgfplots}\usepgfplotslibrary{units}\pgfplotsset{compat=1.16}

\usepackage{setspace}

\usepackage{hyperref}

\graphicspath{{graphics/}{Graphics/}{./}}

\renewenvironment{abstract}{
    \begin{center}
    \textbf{Abstract}
    \vspace{0.5cm}
    \par\itshape
    \begin{minipage}{0.8\linewidth}}{\end{minipage}
    \noindent\ignorespaces
    \end{center}
}

\providecommand{\IEEEkeywordsname}{Keywords}
\newenvironment{IEEEkeywords}
  {\par\small\noindent\textbf{\IEEEkeywordsname---}\ignorespaces}
  {\par\normalsize}
\begin{document}

% =========================
% Front matter
% =========================

\title{Hard Negative Sample--Augmented DPO Post-Training for Small Language Models}

\author{
Haocheng Lu\\ Computer Science, NYU Shanghai\\ \texttt{hl5026@nyu.edu}
\and
Minjun Zhu\\ Computer Science, NYU Shanghai\\ \texttt{mz3487@nyu.edu}
\and
Henry Yu\\ Computer Science, NYU Shanghai\\ \texttt{ky2389@nyu.edu}
}

\maketitle

\begin{abstract}
Large language models (LLMs) continue to struggle with mathematical reasoning, and common post-training pipelines often reduce each generated solution to a binary outcome: correct or incorrect.
This view is limiting in practice, because failures in chain-of-thought (CoT) reasoning are frequently \emph{structured}---solutions may look convincing while containing subtle logical, algebraic, or numerical flaws.
Meanwhile, reinforcement learning from human feedback (RLHF) variants that rely on large reward models or LLM-as-a-judge signals can be expensive, difficult to scale, and unstable to iterate.

We propose a lightweight and pragmatic post-training pipeline that targets such structured errors under realistic compute budgets.
Starting from supervised fine-tuning (SFT) on MetaMathQA-style CoT data, we introduce a compact \emph{MathVerifier} that decomposes a candidate solution into a six-dimensional error profile and aggregates it into interpretable \emph{wrongness} and \emph{absurdity} scores.
These verifier signals serve two roles: (i) mining \emph{hard negatives} that are near-correct yet structurally flawed, and (ii) defining per-sample importance weights that emphasize the most informative preference pairs.
We integrate both into an offline Direct Preference Optimization (DPO) objective via a verifier-guided weighted formulation.

Experiments on a 1.5B-parameter Qwen2.5 model show that verifier-guided, weighted DPO yields more targeted improvements than vanilla SFT and unweighted DPO, particularly on problems where solutions are numerically close to correct but logically inconsistent, while avoiding the overhead of training a large reward model or relying on external judges.
\end{abstract}


% =========================

\section{Introduction}
\label{sec:introduction}

Large language models (LLMs) have demonstrated remarkable progress across diverse tasks, yet mathematical reasoning remains a persistent challenge.
Current post-training and evaluation pipelines overwhelmingly treat each generated solution in a purely binary way---``correct'' or ``incorrect.''
This binary view is fundamentally limiting: in chain-of-thought (CoT) reasoning, failures are frequently \emph{structured}---solutions may appear convincing while containing subtle logical, algebraic, or numerical flaws that a simple correct/incorrect label cannot capture.

The dominant paradigm for providing richer supervision relies on either human annotators or an LLM-as-a-judge, where a stronger model grades candidate responses from the target model.
However, human annotation is prohibitively expensive and slow, especially for mathematical domains requiring specialized expertise.
LLM-as-a-judge approaches, while more scalable, introduce high inference costs, error propagation from the judge model, and systematic biases including self-preference, position bias, and sensitivity to superficial formatting~\cite{krumdick2025nofreelabels,ye2024justice,gu2024surveyllmasjudge}.
These issues are particularly problematic for mathematical reasoning, where correctness is discrete and small logical errors can be masked by fluent, confident text.

To address these limitations, we propose a lightweight and pragmatic post-training pipeline that explicitly targets structured reasoning errors under realistic compute budgets.
Our approach centers on three key ideas: (i)~a compact \emph{MathVerifier} that decomposes candidate solutions into a six-dimensional error profile, enabling fine-grained characterization of \emph{how} a solution is wrong; (ii)~a hard-negative mining strategy that selects solutions which are confidently presented yet structurally or numerically flawed; and (iii)~a verifier-guided weighted Direct Preference Optimization (DPO) objective that emphasizes the most informative preference pairs.
By coupling decomposed verification with hard-negative mining, our framework provides a scalable way to sharpen mathematical reasoning while substantially reducing dependence on human labels and LLM-as-a-judge annotations.

Our main contributions are as follows:
\begin{itemize}
  \item We design a dual-channel \textbf{MathVerifier} that evaluates CoT solutions along six interpretable dimensions and produces aggregate \emph{wrongness} and \emph{absurdity} scores, enabling fine-grained error analysis beyond final-answer correctness.
  \item We propose a \textbf{hard-negative-centered training pipeline} that uses the verifier to mine high-value preference pairs and defines a per-sample importance weight combining verifier error, model confidence, and perplexity, which is integrated into a weighted DPO objective.
  \item We demonstrate that this pipeline yields consistent improvements on GSM8K and MATH benchmarks across three model families at the 1.5B-parameter scale, outperforming both base models and random-uniform DPO, while operating entirely offline and avoiding large reward models or external judges.
\end{itemize}

\section{Related Work}

\paragraph{Verifiers and process supervision for mathematical reasoning.}
A growing line of work improves mathematical reasoning in LLMs by learning explicit verifiers or process reward models.
Lightman et al.\ propose process supervision for mathematical problem solving and introduce the PRM800K dataset, showing that step-level reward models can substantially improve performance on MATH-style benchmarks.~\cite{lightman2023lets}
Building on this idea, Math-Shepherd trains a process reward model from automatically constructed supervision, and uses it both to rerank candidate solutions and as a reward model for step-by-step PPO, yielding large gains on GSM8K and MATH.~\cite{wang2024mathshepherd}

More recently, VeriThinker treats verification itself as an auxiliary task: the model is fine-tuned to judge the correctness of chain-of-thought (CoT) solutions, which leads to shorter reasoning chains with comparable or even higher accuracy.~\cite{chen2025verithinker}
In parallel, DeepSeekMath-V2 integrates self-verification into a specialized mathematical model, emphasizing strict step-wise checking and highlighting the high false-positive rate of current verifiers that often accept logically flawed proofs.~\cite{shao2025deepseekmathv2}
Other work such as MathQ-Verify focuses on verifying the well-posedness of math questions themselves, filtering out inconsistent or ill-formed problems before training.~\cite{shen2025mathqverify}
Compared to these approaches, typically involving heavy-weight PRMs or specialized architectures, our work assumes only a lightweight \emph{math-verifier} and uses it primarily as a tool for mining informative hard negatives rather than as a full-fledged reward model.

\paragraph{LLM-as-a-judge and its limitations.}
Many alignment and evaluation pipelines adopt the \emph{LLM-as-a-judge} paradigm, where a stronger model grades or compares candidate responses from a target model.
This approach has been widely used for automatic evaluation and for constructing preference data at scale, but recent analyzes show that judge models can suffer from substantial biases and instability, including self-preference, position bias, and sensitivity to superficial format differences.~\cite{krumdick2025nofreelabels,ye2024justice,gu2024surveyllmasjudge}
These issues are particularly problematic for mathematical reasoning, where correctness is discrete and small logical errors can be masked by fluent, confident text; empirical studies report high false-positive rates when generic LLM judges are asked to validate math solutions or proofs.
Our framework deliberately avoids using an external LLM judge both for evaluation and for preference-label generation, and instead relies on programmatic checks and a task-specific math-verifier to decide which solutions are plausible yet incorrect.

\paragraph{Preference-based post-training and hard-negative mining.}
Direct Preference Optimization (DPO) has emerged as a simple and effective alternative to traditional RLHF, recasting preference learning as a supervised objective in preference pairs without training an explicit reward model or performing online rollouts.~\cite{rafailov2023dpo}
Subsequent work has proposed numerous variants and extensions that modify the reference policy, regularization, or loss structure, and several studies emphasize that the quality and diversity of preference pairs—especially the presence of strong hard negatives—are often more critical than the choice of objective itself.
In parallel, preference-based methods have been explored in unlearning and safety settings, where negative preference optimization is used to ``push down'' undesired behaviors while maintaining utility on normal data.~\cite{zhang2024npo,fan2024simplicity}
Our work is complementary to these lines: we use standard DPO as the optimization backbone, and focus on an automatic pipeline for constructing high-value preference pairs in the specific context of mathematical reasoning, where the negative samples are confident but structurally or numerically incorrect solutions mined by our verifier.

\paragraph{Semantic similarity, NLI-style reasoning, and ranking-based evaluation.}
To go beyond purely syntactic or final-answer checks, our math-verifier incorporates semantic signals inspired by natural language inference (NLI) and ranking-based evaluation.
NLI corpora such as SNLI and MultiNLI provide large-scale datasets of sentence pairs labeled as entailment, contradiction, or neutral, and have driven progress on models that reason about high-level semantic relations between natural language statements.~\cite{bowman2015snli,williams2018multinli}
In a complementary direction, Wang et al.\ argue that directly using similarity scores as absolute metrics for embeddings can be misleading, and propose EvalRank, a ranking-based intrinsic evaluation method that better correlates with downstream performance.~\cite{wang2022justrank}
We borrow these perspectives by treating candidate solutions as objects to be \emph{ranked} rather than independently scored: our verifier combines programmatic answer checking with semantic consistency and NLI-style signals to rank multiple solutions for the same problem.
This ranking is then used to identify \emph{hard negatives}, responses that are close to the correct solution in surface form or intermediate steps but contain subtle logical or numerical errors.

\section{Methodology}
\label{sec:methodology}

\begin{figure*}[htbp]
  \centering
  \includegraphics[width=\textwidth]{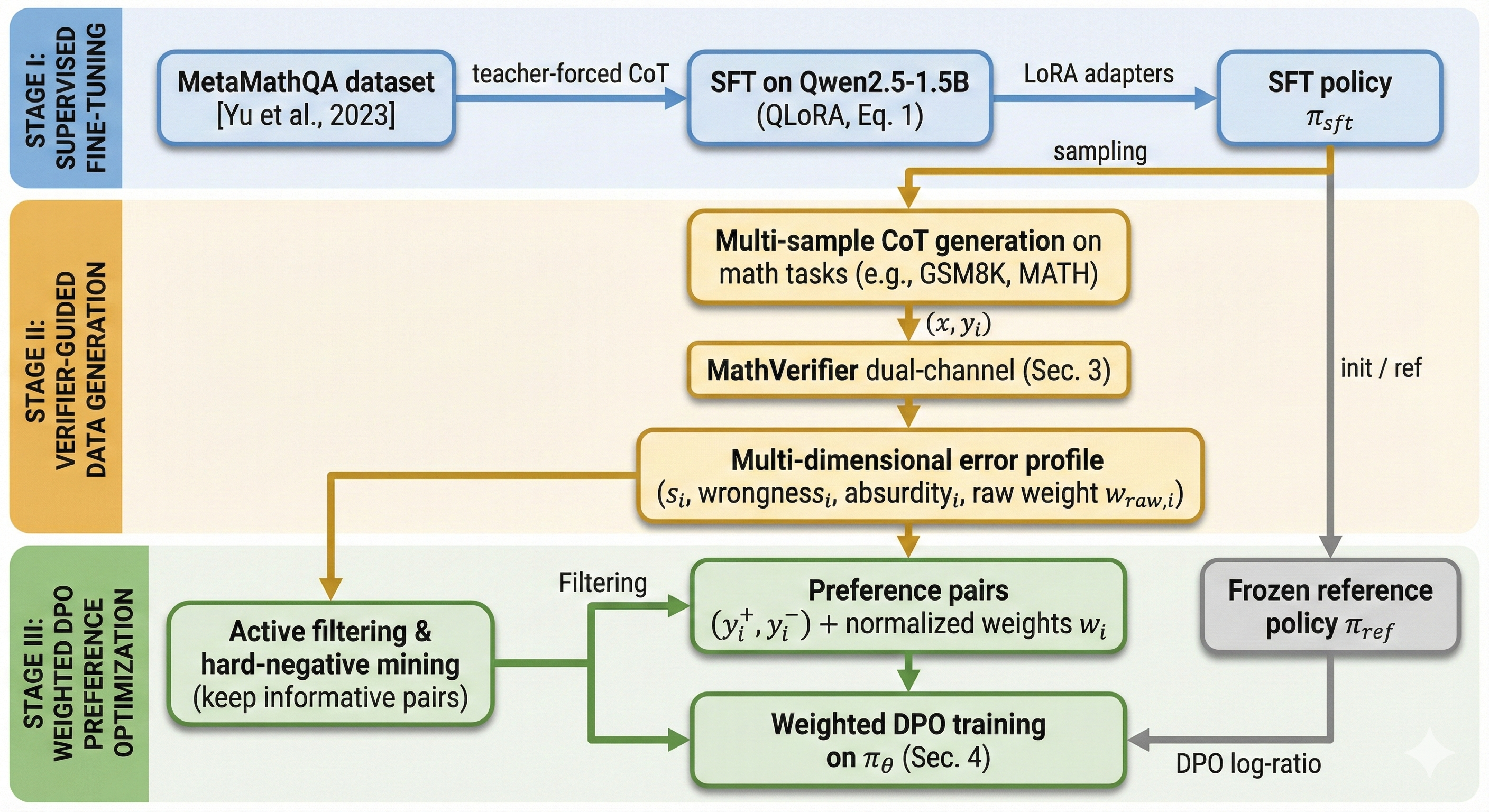}
  \caption{Overall training pipeline. We first perform supervised fine-tuning (SFT) of Qwen2.5-1.5B on MetaMathQA with the standard next-token loss (Eq.~\ref{eq:sft-loss}), obtaining $\pi_{\text{sft}}$. This model is then used to sample multiple chain-of-thought solutions for math benchmarks. The dual-channel MathVerifier assigns each trajectory a multi-dimensional error profile, aggregate wrongness/absurdity scores, and a raw importance weight. These signals drive active filtering and hard-negative mining, yielding weighted preference pairs $(y^+,y^-)$ used in a weighted DPO objective to train the final policy $\pi_\theta$, with a frozen reference policy $\pi_{\text{ref}}$.}
  \label{fig:verifier-dpo-pipeline}
\end{figure*}

In this section, we describe our full training pipeline, which consists of:
(i) supervised fine-tuning on MetaMathQA,
(ii) verifier-guided evaluation of chain-of-thought solutions, and
(iii) verifier-guided hard negative mining with weighted DPO.
Figure~\ref{fig:verifier-dpo-pipeline} provides an overview.

\subsection{Supervised Fine-Tuning on MetaMathQA}
\label{sec:sft-metamath}

Before introducing our verifier-guided training, we first adapt the base
model to math reasoning with standard supervised fine-tuning (SFT) as a cold start for the base model to get familiar with our context of solving math problems.
We use the MetaMathQA dataset~\cite{yu2023metamath,metamathqa-hf},
an AI-augmented corpus of $\sim$395k question–answer pairs derived from the
training splits of GSM8K and MATH. 
Each original math problem is rewritten from multiple perspectives and paired
with a chain-of-thought style solution, without leaking any test-set data
from the underlying benchmarks~\cite{yu2023metamath,metamathqa-hf,metamath-github}. 

The selection of MetaMathQA is motivated by its explicit inclusion of step-by-step reasoning chains, which serves two critical purposes for our pipeline: first, it enables the model to learn structured problem decomposition that is essential for interpretable mathematical reasoning; second, and more importantly, the granular step-level structure provides the necessary foundation for our verifier to identify specific failure modes, enabling targeted hard negative mining in the subsequent DPO stage.  

\paragraph{Model and parameter-efficient setup.}
Our base model is \texttt{Qwen2.5-1.5B-Instruct}. We selected this specific model for three strategic reasons.
First, the Qwen2.5 family provides a unique ecosystem of parallel variants—including general-purpose \emph{Instruct}, specialized \emph{Coder}, and \emph{Math} versions, which allow us to benchmark our self-aligned model against a strong, industry-standard domain expert (\texttt{Qwen2.5-Math-1.5B-Instruct}) within the exact same architecture.
Second, the 1.5B parameter size is the more preferred in our context because it is the only modern architecture small enough to allow our full-pipeline experimentation while still retaining sufficient reasoning capacity, especially given the limited computing power we have.
Third, Qwen2.5 significantly outperforms previous small language models (SLMs) on reasoning benchmarks, providing a non-degenerate starting point where the verifier can actually find meaningful hard negatives rather than just random hallucinations.

We fine-tune it with QLoRA: the backbone weights are loaded in 4-bit NF4 quantization with double quantization to reduce memory requirements, and we insert rank-$16$ LoRA adapters with $\alpha=32$ and dropout $0.05$ targeting all linear projection layers (\texttt{q\_proj}, \texttt{k\_proj}, \texttt{v\_proj}, \texttt{o\_proj}, \texttt{gate\_proj}, \texttt{up\_proj}, \texttt{down\_proj}) in the attention and MLP modules. Only the LoRA parameters are updated; all other weights remain frozen.

\paragraph{Prompt formatting and data packing.}
Each MetaMathQA example provides a \texttt{query} and a CoT-style
\texttt{response}.
We render them into the Qwen chat format via a fixed system prompt
encouraging step-by-step reasoning. The chat template converts the message structure into the Qwen-specific format with special tokens:
\begin{quote}
\small
\texttt{<|im\_start|>system}\\
\texttt{You are a mathematical reasoning expert. Always reason step by step before providing the final answer.<|im\_end|>}\\
\texttt{<|im\_start|>user}\\
\texttt{\{query\}<|im\_end|>}\\
\texttt{<|im\_start|>assistant}\\
\texttt{\{response\}<|im\_end|>}
\end{quote}

We use \texttt{tokenizer.apply\_chat\_template} with \texttt{add\_generation\_prompt=False} to convert this message list into a single token sequence, ensuring the complete conversation including the assistant response is included for training. Sequences are padded on the right to a maximum length of $1024$ tokens and truncated if longer.
The \textsc{SFTTrainer} from TRL~\cite{trl-lib,trl-sft} then applies
length-based packing: multiple formatted conversations are concatenated and
chunked up to the context window, which improves throughput without
changing the underlying objective.

\paragraph{SFT objective and loss masking.}
Let $x = (x_0, \dots, x_{T-1})$ denote a packed training sequence obtained
from the rendered conversations.
The \textsc{SFTTrainer} is a thin wrapper around
\textsc{Trainer} that optimizes the standard auto-regressive
next-token log-likelihood over all non-padding tokens.
Concretely, logits $f_\theta(x_{<t})$ are produced at each position and the
per-token cross-entropy is computed against the shifted labels;
a binary loss mask $m_t \in \{0,1\}$ is used to ignore padding and special
positions, and the loss is
\begin{equation}
\label{eq:sft-loss}
\mathcal{L}_{\text{SFT}}(\theta)
= - \frac{1}{N} \sum_{t=0}^{T-1} m_t
    \log p_\theta(x_t \mid x_{<t}),
\qquad
N = \sum_{t} m_t ,
\end{equation}
which corresponds to the implementation
$\texttt{loss = (per\_token\_loss * loss\_mask).sum() / num\_items}$ in
the TRL codebase~\cite{trl-sft}.
In our setup, we do not apply completion-only masking, so all non-padding
tokens in the rendered conversation (system, user, and assistant turns)
contribute to the loss.
Because assistant tokens constitute the majority of positions, the
effective supervision is dominated by imitating the provided chain-of-thought
solutions under teacher forcing.

\paragraph{Optimization details.}
For the hyperparameters, we use default configurations following standard QLoRA practices. We train for a single epoch over the entire MetaMathQA-395K corpus, with
a global batch size of $32$ (per-device batch size $8$ and gradient
accumulation $4$), learning rate $2\times 10^{-4}$, paged AdamW optimizer,
and gradient checkpointing.
All SFT and DPO experiments are run on a single NVIDIA RTX~5090 GPU
with FlashAttention-2 for efficient attention computation.
SFT training over the full MetaMathQA corpus takes approximately 6~hours of wall-clock time.
The resulting LoRA-adapted model serves as the starting point for the
verifier-guided DPO stage described in Sections~\ref{sec:math-verifier}
and~\ref{sec:verifier-dpo}.

\subsection{Verifier-Guided Evaluation of Mathematical Reasoning}
\label{sec:math-verifier}

Our goal is to evaluate not only whether the final answer is correct, but \emph{how} a solution is wrong: whether the chain-of-thought (CoT) is semantically misaligned, structurally incomplete, logically inconsistent, or merely affected by small arithmetic slips.
We therefore design a lightweight \textbf{MathVerifier} that assigns each solution a multi-dimensional error profile and two aggregate scores: a coarse ``wrongness'' score and a more structural ``absurdity'' score.

\paragraph{Dual-channel architecture.}
The verifier adopts a dual-channel architecture tailored to reinforcement learning and offline post-training.
A \emph{fast channel} builds dense-embedding similarity matrices between the problem statement, a reference solution, and each predicted step, providing cheap estimates of semantic alignment, redundancy, and coverage that can be queried frequently.
A \emph{fine analysis channel} is triggered on sampled candidates and performs more expensive checks, including sequence alignment, NLI-based logical consistency, and symbolic/numerical validation.
Both channels operate on the same CoT but at different computational budgets, and their outputs are combined into a single error profile.

Concretely, we encode steps with \texttt{all-MiniLM-L6-v2}~\cite{reimers2019sentencebert}, a lightweight sentence-transformer that maps each text span to a 384-dimensional dense vector, and compute cosine similarities in batch,
\[
M^{(q)}_{t} = \mathrm{sim}(h(q), h(s_t)), \quad
M^{(r)}_{t} = \mathrm{sim}(h(r_t), h(s_t)),
\]
where $q$ is the question, $r_t$ the $t$-th reference step, and $s_t$ the $t$-th predicted step.
These matrices drive semantic-drift, coverage, and redundancy signals.
In the fine channel, we perform Needleman--Wunsch-style sequence alignment between $\{r_t\}$ and $\{s_t\}$ to detect missing and extra steps, and apply a \texttt{DeBERTa-v3-base} model fine-tuned on MultiNLI~\cite{williams2018multinli} to pairs $(s_{t-1}, s_t)$ and $(q, s_t)$ to detect contradictions and non-sequiturs.

\paragraph{Six-dimensional evaluation.}
On top of this architecture, the verifier instantiates a six-dimensional evaluation framework; each solution is mapped to
\[
\mathbf{s} = (s_{\text{sem}}, s_{\text{struct}}, s_{\text{order}},
             s_{\text{logic}}, s_{\text{sym}}, s_{\text{ans}}),
\]
where each component measures a specific error type:

\begin{enumerate}
  \item \textbf{Semantic content difference} $s_{\text{sem}}$ measures alignment between predicted and reference steps.
        It increases under semantic drift, irrelevant content, or mismatched operations, and is near zero for paraphrased but equivalent reasoning.
  \item \textbf{Step structural integrity} $s_{\text{struct}}$ evaluates coverage and redundancy.
        Sequence alignment yields unmatched reference steps (missing reasoning) and unmatched prediction steps (redundant or tangential reasoning), which are converted into a structural penalty.
  \item \textbf{Step order preservation} $s_{\text{order}}$ measures whether the reasoning follows a plausible progression (e.g., define variables $\rightarrow$ form equations $\rightarrow$ solve $\rightarrow$ verify).
        We treat aligned indices as two permutations and use rank-based statistics (e.g., Spearman correlation), plus penalties for inversions of key steps.
  \item \textbf{Logical inference consistency} $s_{\text{logic}}$ uses an NLI head to detect contradictions and non-sequiturs between successive steps and between steps and given conditions.
  \item \textbf{Numerical vs.\ symbolic error analysis} $s_{\text{sym}}$ uses a symbolic engine (e.g., SymPy) to normalize expressions and distinguish arithmetic slips from conceptual or formula-level errors.
        Wrong formulas (e.g., using area for perimeter) incur high symbolic penalties, whereas small arithmetic mistakes are treated as lighter numerical errors.
  \item \textbf{Final answer consistency} $s_{\text{ans}}$ checks whether the final answer is consistent with the reference answer under exact or relaxed equivalence.
\end{enumerate}

This decomposition lets us separate, for example, ``numerically wrong but structurally correct'' solutions from ``numerically correct but conceptually invalid'' ones, which is crucial for constructing informative hard negatives.

\paragraph{Absurdity and wrongness scores.}
The six-dimensional vector is further collapsed into two scalar scores.
First, we define an overall \emph{wrongness} score
\[
\text{wrongness} = \sum_{k} w_k \, s_k,
\]
where $w_k$ are empirically chosen, task-dependent weights over the six dimensions.
In all experiments we use
$w_{\text{ans}} = 0.25$,
$w_{\text{logic}} = 0.20$,
$w_{\text{sym}} = 0.15$,
$w_{\text{sem}} = 0.15$,
$w_{\text{struct}} = 0.15$,
$w_{\text{order}} = 0.10$,
reflecting the intuition that final-answer correctness and logical consistency are the strongest indicators of overall solution quality.
Wrongness is used as a coarse reward-shaping and ranking signal.

Second, we define an \emph{absurdity} score that emphasizes structural and logical pathologies:
\[
\text{absurdity} = \alpha_{\text{logic}} s_{\text{logic}}
                 + \alpha_{\text{struct}} s_{\text{struct}}
                 + \alpha_{\text{order}} s_{\text{order}}
                 + \alpha_{\text{sem}} s_{\text{sem}},
\]
with $\alpha_{\text{logic}} = 0.35$, $\alpha_{\text{struct}} = 0.30$, $\alpha_{\text{order}} = 0.20$, $\alpha_{\text{sem}} = 0.15$,
deliberately downweighting purely numerical or final-answer discrepancies.
Intuitively, absurdity measures how ``structurally unreasonable'' a solution is, even when the final scalar happens to match the correct answer.

\paragraph{Per-sample importance score.}
In addition to the error profile, we associate each solution with a raw \emph{dataset weight} that reflects how informative it is for preference learning.
For a candidate solution $y_i$ we define
\begin{equation}
  w_{\text{raw}, i}
  = \text{wrongness}_i
  + \bigl(1 - c_i\bigr)
  + \frac{p_i}{100},
  \label{eq:mathverifier-raw-weight}
\end{equation}

where $c_i \in [0,1]$ is the mean token-level probability of the final-answer span under the current policy, smoothed with a sigmoid transformation $c_i = \sigma\!\bigl(\bar{\log p}_{\text{ans},i}\bigr)$ to ensure a well-behaved range, and $p_i$ is the sequence-level perplexity of the full CoT trajectory under the same policy.
The division by $100$ is a normalizing constant chosen so that $p_i/100$ has a comparable magnitude to the other two terms, given that perplexity values in our setting typically range from $5$ to $150$.
Samples that are structurally problematic (high wrongness), unexpectedly low-confidence, or assigned high perplexity receive larger $w_{\text{raw}, i}$ and are considered more valuable for training.
These raw weights are later normalized and clipped at the batch level in our weighted DPO objective (Sec.~\ref{sec:verifier-dpo}), but their definition is fully determined by verifier signals and pre-existing model statistics.

\paragraph{Verifier-guided hard negative mining.}
Given a question, we sample multiple CoT solutions from the base model and score each with the verifier, obtaining $(\text{wrongness}, \text{absurdity}, \mathbf{s}, w_{\text{raw}})$.
We then select \emph{hard negatives} as solutions that are
(i) high-confidence according to the model,
(ii) near the decision boundary in terms of final-answer correctness or semantic proximity to a good solution, and
(iii) assigned relatively high absurdity and/or specific structural or logical error tags.
Because the verifier exposes rich, interpretable error dimensions, the resulting hard negatives cover diverse failure modes (semantic drift, missing steps, redundant or reordered reasoning, symbolic misuse), making the preference signal more informative than a purely outcome-based correct/incorrect label.

\paragraph{Efficiency considerations.}
To make verifier-guided training tractable, we cache embedding and NLI calls across repeated patterns, compute similarity matrices in batches with normalized cosine scores, and restrict expensive operations (alignment, NLI, symbolic checks) to a fine analysis pass on a subset of candidates.
After these optimizations, the average per-sample evaluation time is reduced from $\approx 4.6$\,s to $\approx 0.6$\,s on CPU, enabling integration of the verifier into large-scale RL and DPO pipelines.

\subsection{Verifier-Guided Hard Negative Mining and Weighted DPO}
\label{sec:verifier-dpo}

Given the multi-dimensional error profile produced by the MathVerifier,
\((\text{wrongness}, \text{absurdity}, \mathbf{s})\),
we use the verifier to (i) actively filter a large pool of model-generated solutions
into a smaller, high-value subset, (ii) select hard negatives, and (iii) train the
policy with a weighted DPO objective in an entirely offline manner.

\paragraph{Active filtering at the dataset level.}
For each problem $x$ in a math dataset, we first sample multiple chain-of-thought
solutions $\{y_i\}$ from a base model (including correct, incorrect, and partially
correct trajectories).
The MathVerifier is then applied to every $(x, y_i)$, yielding
\[
\text{wrongness}_i,\quad
\text{absurdity}_i,\quad
\mathbf{s}_i,\quad
\text{primary\_error}_i.
\]
We treat the verifier as an active-learning style filter:
trivial samples (either almost perfect or completely degenerate) are discarded
or heavily downweighted, while we retain solutions where the model is reasonably
confident, the answer or structure is close to correct, but the verifier reveals
clear structural or logical flaws.
The result is a \emph{learnable subset} with much higher information density
than the raw pool of completions.

\paragraph{Verifier-guided hard negatives.}
Within this learnable subset, we use the verifier to select \emph{hard negatives}.
Intuitively, these are solutions that the model takes seriously, that look similar
to a correct reasoning path, but that contain decisive structural or logical errors.
Concretely, a candidate $y_i$ is considered a hard negative if it satisfies:
(i) high model confidence (e.g., large answer probability or high step-level
log-probability);
(ii) proximity to a good solution in terms of final answer or semantic alignment
(e.g., near-miss answers or CoTs that align well with a reference solution except for
one or two crucial steps); and
(iii) medium-to-high $\text{wrongness}_i$ and elevated $\text{absurdity}_i$, or
clear failures in structural/logical dimensions of $\mathbf{s}_i$
(e.g., missing steps, redundancy, order errors, logical flips).
Such samples force the model to distinguish truly valid reasoning from
``almost correct but fatally flawed'' trajectories, providing a much richer
preference signal than a simple correct/incorrect label.

\paragraph{Preference pairs and DPO objective.}
For each problem $x$, we construct a preference pair $(y^+, y^-)$:
$y^+$ is a correct and structurally coherent solution or, more generally,
one that is clearly preferred by the verifier; $y^-$ is a verifier-identified
hard negative.
We then apply the standard DPO per-sample loss
\begin{equation}
\ell_i^{\text{DPO}}
= - \log \sigma\Big(
  \beta\big[
    \log \pi_\theta(y^+ \mid x)
    - \log \pi_\theta(y^- \mid x)
    - \log \pi_{\text{ref}}(y^+ \mid x)
    + \log \pi_{\text{ref}}(y^- \mid x)
  \big]
\Big)
\label{eq:dpo-loss}
\end{equation}

where $\pi_\theta$ is the trainable policy, $\pi_{\text{ref}}$ is the frozen SFT checkpoint from Section~\ref{sec:sft-metamath} serving as the reference policy, and $\beta = 0.1$ is the inverse temperature controlling the strength of the KL penalty, following the default value from Rafailov et al.~\cite{rafailov2023dpo}.

\paragraph{Raw sample weight from verifier and model signals.}
Not all preference pairs are equally informative.
We therefore assign each pair a \emph{raw} dataset weight $w_{\text{raw},i}$ that
combines verifier scores with existing model diagnostics.
Let $\text{wrongness}_i$ be the verifier score defined in
Section~\ref{sec:math-verifier},
$\text{conf}_i \in [0,1]$ be an existing confidence estimate for $(x,y_i)$
(e.g., the calibrated probability of the final answer),
and $\text{ppl}_i$ be the perplexity of the solution under the base model.
We define the raw per-example weight $w_{\text{raw},i}$ as in
Eq.~\eqref{eq:mathverifier-raw-weight}.
This weighting scheme upweights examples that are structurally problematic
(high $\text{wrongness}_i$), under-confident or uncertain
(large $1-\text{conf}_i$), and linguistically or probabilistically challenging
(high perplexity), while keeping all three terms on a comparable scale via
the normalization factor $100$.
In practice, $w_{\text{raw},i}$ serves as a coarse estimate of how much
additional signal a given preference pair can provide during training.

\paragraph{Batch-wise normalization and safe clipping.}
To avoid destabilizing optimization, we normalize these weights \emph{within
each batch} through three steps, which is fully compatible with standard
DPO trainers.

\emph{(1) Normalization (unit mean).}
For a batch of size $B$, we first enforce unit mean:
\begin{equation}
w_{\text{norm}, i}
= \frac{
    w_{\text{raw}, i}
}{
    \frac{1}{B} \sum_{j=1}^{B} w_{\text{raw}, j}
}
\label{eq:weight-normalization}
\end{equation}

This keeps the effective learning rate unchanged at the batch level.

\emph{(2) $\lambda$-interpolation (tempering the strength).}
We then pull the normalized weights towards $1$ using a scalar
$\lambda \in [0,1]$:
\begin{equation}
w_{\text{eff}, i}
= 1 + \lambda \bigl(w_{\text{norm}, i} - 1\bigr)
\label{eq:weight-effective}
\end{equation}

When $\lambda = 0$, dataset weights are ignored (all samples are equally weighted);
when $\lambda = 1$, we use full-strength weights.
In practice, we fix $\lambda = 0.3$ (i.e., \texttt{dataset\_weight\_lambda = 0.3}
in our implementation), which preserves ``who is heavier/lighter'' information
without making training brittle.

\emph{(3) Safe clipping.}
Finally, we apply a safety clip to prevent extreme examples from dominating:
\begin{equation}
w_i
= \mathrm{clip}\bigl(w_{\text{eff}, i},\, w_{\min},\, w_{\max}\bigr),
\label{eq:weight-clipped}
\end{equation}

and in all experiments we set $w_{\min} = 0.5$ and $w_{\max} = 2.0$
(corresponding to \texttt{dataset\_weight\_min = 0.5} and
\texttt{dataset\_weight\_max = 2.0}), so that even the most important
samples are at most about twice as heavy as the average, and the least
informative samples still contribute some gradient.

The batch loss is then
\begin{equation}
L
= \frac{1}{B} \sum_{i=1}^{B} w_i \, \ell_i^{\text{DPO}}
\label{eq:dpo-weighted-loss}
\end{equation}

Overall, the MathVerifier provides an active-learning style filter and a
sample-importance signal at the dataset level.
The weighted DPO objective then smooths these weights within each batch
to maintain optimization stability.
Together, they yield an offline preference-learning pipeline that
systematically focuses training on verifier-identified hard negatives
covering diverse failure modes (semantic drift, missing steps, redundancy,
order errors, symbolic misuse, \emph{etc.}).

\paragraph{Filtering pipeline and data statistics.}
Starting from an offline pool of approximately $6\times 10^5$ DeepSeek-R1-style CoT trajectories derived from the HuggingFaceH4/numina-deepseek-r1-qwen-7b dataset~\cite{huggingfaceh4-numina-deepseek-r1-qwen-7b}, the verifier applies a multi-stage filtering pipeline:
\begin{enumerate}
  \item \emph{Degenerate removal:} Empty, truncated, or nonsensical outputs are discarded, retaining $\approx$520K trajectories.
  \item \emph{Trivially correct removal:} Solutions where the final answer is correct and all six verifier dimensions score below $0.05$ are removed as uninformative positives, leaving $\approx$380K.
  \item \emph{Trivially wrong removal:} Solutions with wrongness $> 0.9$ and absurdity $> 0.8$ (completely off-track responses) are discarded, leaving $\approx$280K.
  \item \emph{Hard-negative selection and pair construction:} From the remaining pool, we select preference pairs where the negative candidate satisfies the hard-negative criteria (moderate wrongness in $[0.3, 0.7]$ and at least one elevated structural or logical dimension), yielding approximately $10^4$ high-value preference pairs.
\end{enumerate}
This four-stage process reduces the data by roughly $98\%$, concentrating the training signal on the most informative examples.

\subsection{Evaluation implementation}
\label{sec:eval-impl}

For the actual evaluation on GSM8K and MATH, we follow the standard
\emph{Step-DPO} evaluation setup and adapt their open-source script
\texttt{eval\_math.py} from the official repository~\cite{lai2024stepdpo,stepdpo-github}.

Concretely, the script takes as input a model checkpoint, a JSONL test
file (either GSM8K or MATH), a prompt template name (e.g., \texttt{qwen2-boxed}),
and an output path.
For each test problem it:
\begin{enumerate}
  \item wraps the question into a model-specific prompt (matching the
        \texttt{qwen2-boxed} format used in Step-DPO),
  \item runs batched generation with greedy decoding (temperature~$=0$),
        using the model in HF \texttt{AutoModelForCausalLM} format,
  \item extracts the final boxed answer from the generated chain-of-thought
        using the same regular-expression based parser as in the original
        script,
  \item normalizes both prediction and reference (e.g., stripping spaces,
        commas, and trivial unit tokens), and
  \item computes exact match accuracy and saves per-sample predictions
        and correctness flags to a JSON file for later analysis.
\end{enumerate}

We keep these evaluation settings unchanged across all model variants
(base, SFT-only, and verifier-guided DPO), so that the accuracy numbers
reported in Table~\ref{tab:gsm8k-main} are directly comparable to each
other and to prior Step-DPO results under the same GSM8K/MATH evaluation
protocol.
The full evaluation script is available at the Step-DPO GitHub repository
(\texttt{eval\_math.py})~\cite{stepdpo-github}.

\section{Experiments}
\label{sec:experiments}

\subsection{Experimentation protocol}
\label{sec:exp-protocol}

Our experimental goal is to isolate the effect of verifier-guided hard-negative
selection and weighting, rather than the effect of supervised pre-training
itself.
All models therefore start from the same SFT checkpoint described in
Section~\ref{sec:sft-metamath} (Qwen2.5-1.5B-Instruct fine-tuned on
MetaMathQA), and we compare different DPO configurations on top of this
shared baseline.
The offline preference pool is constructed as described in
Sec.~\ref{sec:math-verifier} from trajectories generated on top of the
\texttt{HuggingFaceH4/numina-deepseek-r1-qwen-7b} dataset~\cite{huggingfaceh4-numina-deepseek-r1-qwen-7b},
before and after the MathVerifier-based filtering.

\paragraph{Training conditions.}
For each training problem $x$ in our offline pool, we first sample multiple
chain-of-thought solutions from the SFT model.
From this common candidate set, we construct two DPO datasets:
\begin{itemize}
  \item \textbf{Random-uniform DPO (baseline).}
        We randomly sample preference pairs $(y^+, y^-)$ from the pool
        using only final-answer correctness to define preferences, and we
        assign each pair a uniform weight $w_i = 1$ (i.e., no verifier
        scores and no importance re-weighting).
  \item \textbf{Verifier-guided weighted DPO (ours).}
        We apply the MathVerifier to all candidates, select hard negatives
        according to the criteria in Sec.~\ref{sec:verifier-dpo}, and
        assign each preference pair a per-sample importance weight $w_i$
        using the weighting scheme defined in Sec.~\ref{sec:math-verifier}
        (based on the verifier wrongness score, model confidence, and
        trajectory perplexity), followed by batch-level normalization and
        clipping with hyperparameters
        $\lambda = 0.3$, $w_{\min} = 0.5$, $w_{\max} = 2.0$.
\end{itemize}
In both settings we keep the number of DPO steps, batch size, learning
rate, and all other optimization hyperparameters identical, so that any
difference in downstream performance can be attributed to the way
preference data are selected and weighted.
All DPO runs are trained on a single NVIDIA RTX~5090 GPU; one full run
over the roughly $10^4$ verifier-selected preference pairs takes about
4 hours of wall-clock time.

\paragraph{Evaluation metrics.}
We follow the Step-DPO evaluation protocol for GSM8K and MATH: exact match
accuracy on the official test sets, using a shared decoding setup and
answer-extraction procedure for all model variants
(see Sec.~\ref{sec:eval-impl} for implementation details).
All numbers reported in this section are computed under this protocol and
differ only in the DPO configuration applied on top of the same SFT checkpoint.
MATH results are reported on the benchmark of Hendrycks et al.~\cite{hendrycksmath2021}.

\begin{figure}[htbp]
  \centering
  \includegraphics[width=0.8\linewidth]{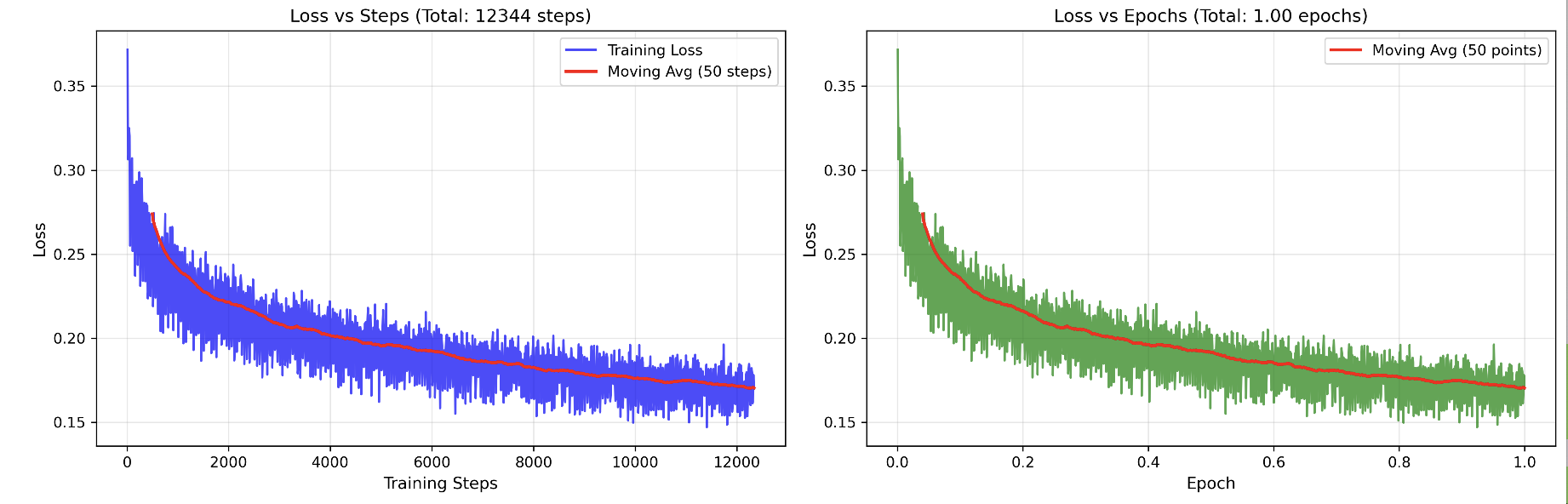}
  \caption{Training curves of the SFT base model used for all experiments
  (Qwen2.5-1.5B-Instruct fine-tuned on MetaMathQA).}
  \label{fig:sft-training}
\end{figure}

\subsection{Main GSM8K and MATH accuracy results}
\label{sec:gsm8k-results}

Table~\ref{tab:gsm8k-main} summarizes the exact match accuracy on GSM8K
and MATH for three model families and for three DPO configurations
per family: the base model (no DPO), random-uniform DPO, and our
verifier-guided hard-negative DPO.
For each family, we always start from the same base checkpoint and only
change how preference data are constructed and weighted.

\begin{figure}[t]
  \centering
  \includegraphics[width=0.7\linewidth]{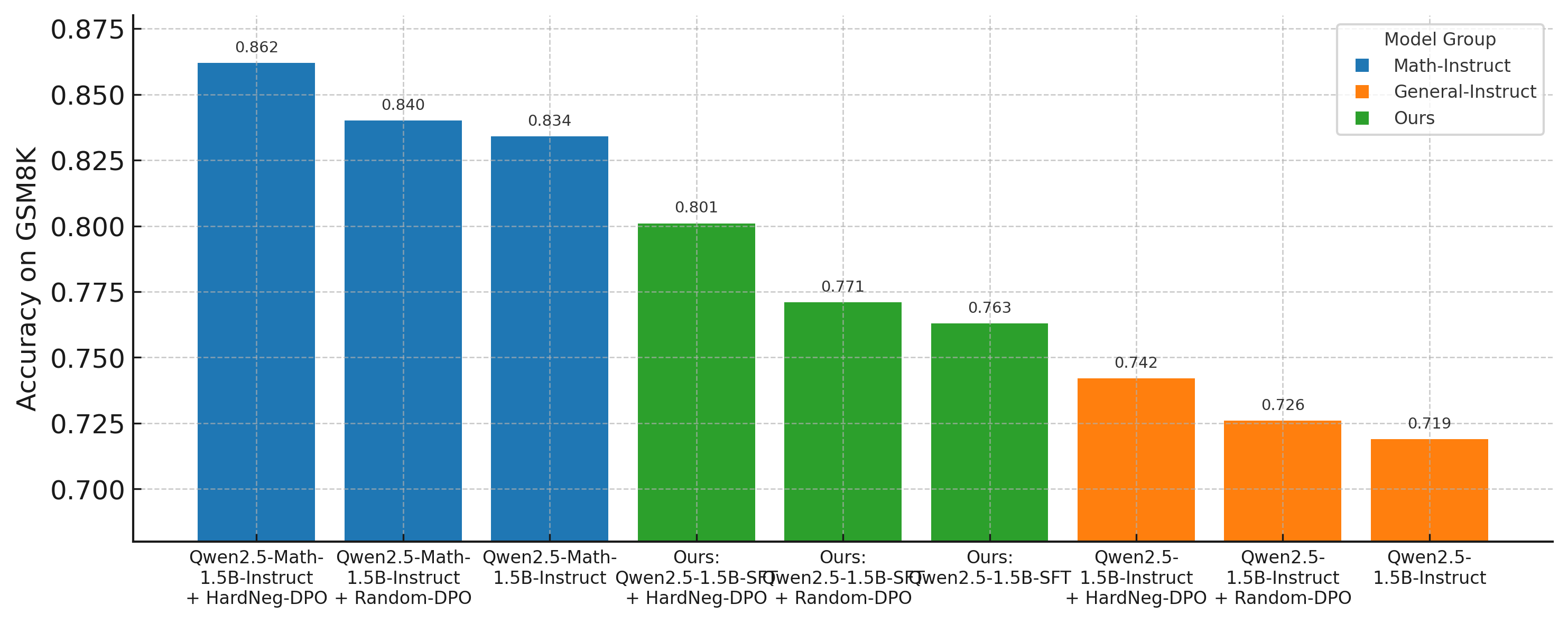}
  \caption{GSM8K exact match accuracy for the SFT baseline, random-uniform
  DPO, and verifier-guided hard-negative DPO across three model families.
  Bars show mean accuracy over seeds (five seeds for random DPO, three
  for hard-neg DPO).
  In every case, hard-negative DPO achieves higher accuracy than the
  corresponding random-DPO baseline on top of the same base model.}
  \label{fig:gsm8k-acc}
\end{figure}

\begin{figure}[htbp]
  \centering
  \includegraphics[width=0.7\linewidth]{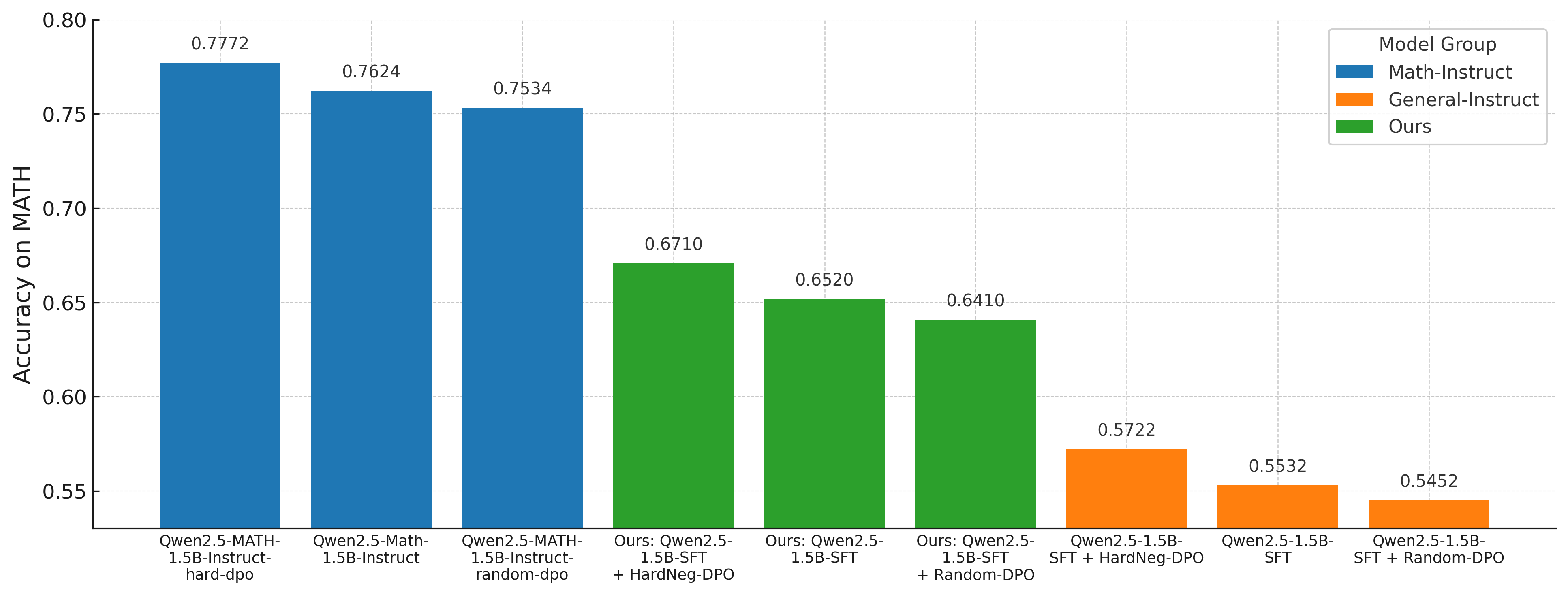}
  \caption{MATH exact match accuracy for the same set of models as in
  Table~\ref{tab:gsm8k-main}.
  As on GSM8K, verifier-guided hard-negative DPO consistently improves
  over the corresponding base model and typically outperforms random-uniform
  DPO, although random DPO can slightly degrade MATH performance in some
  cases.}
  \label{fig:math-acc}
\end{figure}

To reduce variance from stochastic sampling, the random-uniform DPO
configuration is run with five different random seeds and the verifier-guided
configuration with three seeds; we report mean $\pm$ standard deviation for
all multi-seed runs.
Figure~\ref{fig:gsm8k-acc} and Figure~\ref{fig:math-acc} visualize the
same numbers for GSM8K and MATH, respectively.

\begin{table}[t]
  \centering
  \small
  \begin{tabular}{lccc}
    \toprule
    Model (1.5B parameters) & MATH & GSM8K & Remark \\
    \midrule
    Qwen2.5-1.5B-Instruct                    & 55.3 & 71.9 & base \\
    Qwen2.5-1.5B-Instruct + Random DPO       & 54.5{\scriptsize$\pm$0.6} & 72.6{\scriptsize$\pm$0.4} & random-uniform \\
    Qwen2.5-1.5B-Instruct + Hard-neg DPO     & \textbf{57.2}{\scriptsize$\pm$0.4} & \textbf{74.2}{\scriptsize$\pm$0.3} & ours \\
    \midrule
    Qwen2.5-1.5B-SFT (MetaMathQA, ours)      & 65.2 & 76.3 & base SFT \\
    Qwen2.5-1.5B-SFT + Random DPO            & 64.1{\scriptsize$\pm$0.7} & 77.1{\scriptsize$\pm$0.5} & random-uniform \\
    Qwen2.5-1.5B-SFT + Hard-neg DPO          & \textbf{67.1}{\scriptsize$\pm$0.5} & \textbf{80.1}{\scriptsize$\pm$0.4} & ours \\
    \midrule
    Qwen2.5-Math-1.5B-Instruct               & 76.2 & 83.4 & base \\
    Qwen2.5-Math-1.5B-Instruct + Random DPO  & 75.3{\scriptsize$\pm$0.5} & 84.0{\scriptsize$\pm$0.4} & random-uniform \\
    Qwen2.5-Math-1.5B-Instruct + Hard-neg DPO & \textbf{77.7}{\scriptsize$\pm$0.4} & \textbf{86.2}{\scriptsize$\pm$0.3} & ours \\
    \bottomrule
  \end{tabular}
  \caption{Joint MATH and GSM8K exact match accuracy (\%) for three
  model families and three DPO configurations.
  Multi-seed runs report mean $\pm$ standard deviation (five seeds for
  random-uniform DPO, three for hard-negative DPO).
  In each block, the base row reports the performance of the underlying
  checkpoint without DPO, while the following rows report the effect of
  applying random-uniform or verifier-guided hard-negative DPO on top.
  Numbers in parentheses denote absolute gains or drops relative to the
  corresponding base row.}
  \label{tab:gsm8k-main}
\end{table}

Across both datasets, a consistent pattern emerges.
Random-uniform DPO provides a small but stable improvement on GSM8K
($\approx 0.6$--$0.8$ absolute points over the respective base models),
but its effect on MATH is mixed: it slightly improves the Qwen2.5-1.5B
instruction-tuned model, yet degrades performance for the SFT and
Qwen2.5-Math-1.5B families.
In contrast, our verifier-guided hard-negative DPO configuration yields
robust gains on both benchmarks.
On GSM8K, hard-negative DPO improves over the corresponding base models
by about $2.3$--$3.8$ points and over the random-DPO baselines by about
$2$ points.
On MATH, it consistently adds $1.5$--$1.9$ points over the base models and
reverses the degradation introduced by random-uniform DPO.
These trends support the hypothesis that the main benefit comes from
focusing optimization on verifier-selected hard negatives rather than
from simply adding more preference data.

\paragraph{Why does random DPO degrade MATH performance?}
A notable pattern in Table~\ref{tab:gsm8k-main} is that random-uniform DPO consistently \emph{hurts} MATH accuracy ($-0.8$ to $-1.1$ points) while slightly helping GSM8K.
We attribute this to the higher complexity and topical diversity of MATH compared to GSM8K: MATH spans algebra, number theory, geometry, combinatorics, and precalculus, and its problems often require multi-step reasoning where the \emph{path} matters more than the final answer.
Random preference pairs based solely on final-answer correctness can be noisy in this regime: a ``correct'' solution may have reached the right answer through flawed reasoning (a false positive), while an ``incorrect'' solution may contain sound intermediate steps derailed by a single arithmetic error.
Training on such noisy pairs can push the model away from valid reasoning strategies---an effect sometimes called the \emph{alignment tax}~\cite{rafailov2023dpo}.
GSM8K problems, being simpler and more formulaic, are less susceptible to this noise because the correct/incorrect signal is more reliable.
Verifier-guided DPO avoids this degradation precisely because the verifier filters out such noisy pairs and concentrates on structurally informative hard negatives.

\subsection{Qualitative analysis: anatomy of a hard negative}
\label{sec:case-study}

To illustrate what a verifier-identified hard negative looks like in practice, Table~\ref{tab:case-study} presents a representative example from the curated preference set.

\begin{table}[htbp]
  \centering
  \small
  \begin{tabular}{p{0.12\linewidth} p{0.80\linewidth}}
    \toprule
    \multicolumn{2}{l}{\textbf{Problem:} John buys 4 notebooks at \$3 each and receives a 10\% discount on the total.} \\
    \multicolumn{2}{l}{He pays with a \$20 bill. How much change does he receive?} \\
    \midrule
    \textbf{$y^+$} (correct) &
    Step 1: Total before discount $= 4 \times 3 = 12$. \newline
    Step 2: Discount $= 0.10 \times 12 = 1.20$. \newline
    Step 3: Total after discount $= 12 - 1.20 = 10.80$. \newline
    Step 4: Change $= 20 - 10.80 = \boxed{9.20}$. \\
    \midrule
    \textbf{$y^-$} (hard neg.) &
    Step 1: Total before discount $= 4 \times 3 = 12$. \newline
    Step 2: Discount per notebook $= 0.10 \times 3 = 0.30$. \newline
    Step 3: Total discount $= 4 \times 0.30 = 1.20$. \newline
    Step 4: Total after discount $= 12 - 1.20 = 10.80$. \newline
    Step 5: Change $= 20 - 12 = \boxed{8.00}$. \\
    \midrule
    \textbf{Verifier} &
    $s_{\text{sem}}\!=\!0.08$,\; $s_{\text{struct}}\!=\!0.05$,\; $s_{\text{order}}\!=\!0.02$,\; $s_{\text{logic}}\!=\!0.41$,\; $s_{\text{sym}}\!=\!0.12$,\; $s_{\text{ans}}\!=\!1.0$. \newline
    The NLI channel flags a \emph{contradiction} between Step~4 (computes \$10.80) and Step~5 (subtracts \$12 instead), yielding elevated $s_{\text{logic}}$.
    All other dimensions remain low because the intermediate reasoning is largely correct. \\
    \bottomrule
  \end{tabular}
  \caption{A representative hard-negative example. The rejected solution $y^-$ correctly computes the discounted total (\$10.80) through Steps~1--4, but reverts to the pre-discount price in the final subtraction (Step~5). This subtle inconsistency is caught by the verifier's logical-inference dimension ($s_{\text{logic}} = 0.41$), while the structural and semantic dimensions remain low, confirming that the error is localized rather than systemic.}
  \label{tab:case-study}
\end{table}

This example highlights the value of decomposed verification: a purely outcome-based label would mark $y^-$ as simply ``wrong,'' whereas the verifier pinpoints the failure to a single logical inconsistency in an otherwise sound chain of reasoning---exactly the type of structured error that makes $y^-$ an informative training signal for DPO.

\subsection{Training dynamics and verifier behavior}
\label{sec:training-dynamics}

Beyond final accuracy, we also inspect the training dynamics of SFT and
DPO, as well as the reward distribution induced by the MathVerifier.

Figure~\ref{fig:sft-training} shows the SFT loss curves on MetaMathQA:
the model converges smoothly over one epoch, providing a stable starting
point for all subsequent DPO experiments.

\begin{figure}[htbp]
  \centering
  \captionsetup[subfigure]{justification=centering}

  \begin{subfigure}[t]{0.48\linewidth}
    \centering
    \includegraphics[height=0.33\textwidth]{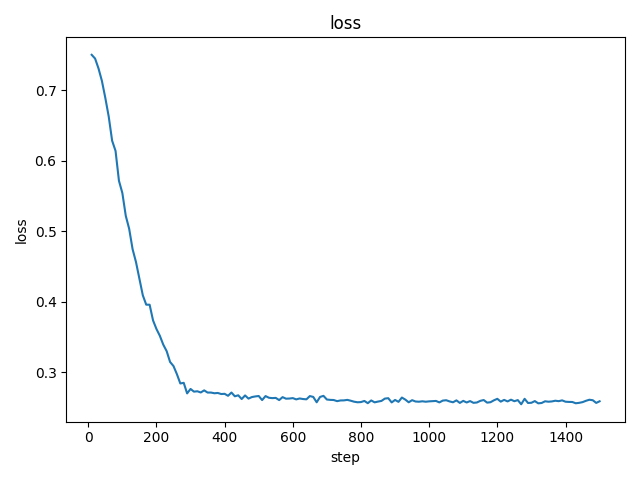}
    \caption{DPO training loss over optimization steps for the random-uniform
    and verifier-guided hard-negative DPO configurations.}
    \label{fig:dpo-loss}
  \end{subfigure}\hfill
  \begin{subfigure}[t]{0.48\linewidth}
    \centering
    \includegraphics[height=0.33\textwidth]{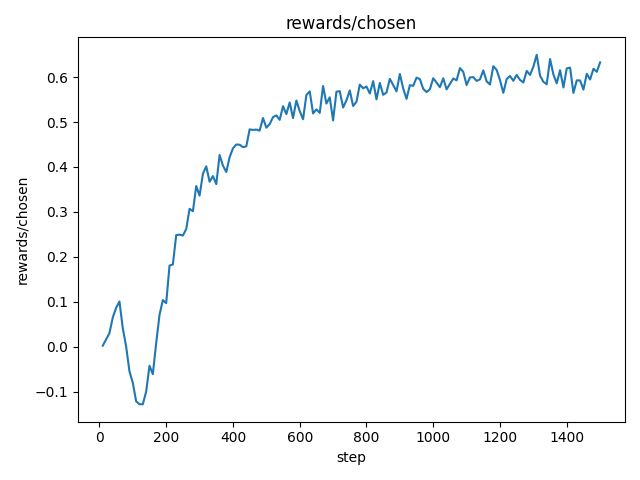}
    \caption{Verifier reward statistics on the curated DPO preference pairs,
    illustrating how MathVerifier concentrates probability mass on
    diverse, high-signal hard negatives.}
    \label{fig:reward-stats}
  \end{subfigure}

  \caption{Training dynamics of DPO optimization and MathVerifier-guided
  reward shaping on the curated preference dataset.}
  \label{fig:dpo-dynamics}
\end{figure}

Figures~\ref{fig:dpo-dynamics} further examine the DPO stage.
Panel~\ref{fig:dpo-loss} plots the training loss versus optimization
steps for both the random-uniform and the verifier-guided hard-negative
DPO configurations.
The overall convergence trends are similar, but the verifier-guided run
reaches a slightly lower loss, consistent with the accuracy improvements
reported in Table~\ref{tab:gsm8k-main}.
Panel~\ref{fig:reward-stats} shows statistics of the verifier-derived
rewards on the curated preference pairs: compared to the random baseline,
MathVerifier concentrates probability mass on diverse, high-signal hard
negatives rather than on trivial correct/incorrect pairs.
This supports our interpretation that the main benefit of the pipeline
lies not in changing the optimization algorithm, but in reshaping which
examples the model learns the most from.

\section{Discussion}
\paragraph{What is different about our pipeline.}
Compared to prior work that either trains large process reward models or
relies on LLM-as-a-judge for preference data and evaluation, the main
novelty of our approach is \emph{how} we organize the training pipeline.
We place hard negatives---solutions that look convincing but are
structurally wrong---at the center of the design, and support them with a
lightweight, decomposed MathVerifier.
The verifier provides a six-dimensional error profile and aggregate
wrongness/absurdity scores, which makes it straightforward to detect
``almost correct'' but logically flawed chains-of-thought and to use them
as the primary negative signal in DPO, rather than treating them as a
byproduct of training.

On the learning side, we move away from purely unweighted preference
optimization and derive a simple per-sample importance score
$w_{\text{raw},i}$ as in Eq.~\eqref{eq:mathverifier-raw-weight},
where $\text{wrongness}_i$ is the verifier's aggregate error,
$c_i$ is an existing confidence estimate, and $p_i$ is the trajectory
perplexity.
After batch-wise normalization and clipping
(Sec.~\ref{sec:verifier-dpo}), this weight can be plugged into the DPO
loss without changing the overall training recipe, but it biases learning
towards structurally problematic, uncertain, and genuinely hard samples.

\paragraph{Distribution mismatch between teacher and student.}
An important design consideration is that our offline preference pool is generated by a 7B-parameter model (DeepSeek-R1-Qwen-7B), while the policy being trained is a 1.5B model.
This creates an off-policy distribution mismatch: the CoT trajectories in the preference data may contain reasoning patterns or vocabulary distributions that the smaller model would not naturally produce.
In principle, DPO is more robust to such mismatch than on-policy methods like PPO, because it operates on fixed preference pairs rather than requiring the policy to generate its own rollouts.
Empirically, we observe that this mismatch does not prevent learning---the 1.5B model still benefits from the curated hard negatives---but it may limit the ceiling of improvement, as some preference pairs may be out of distribution for the smaller model.
A natural extension would be to generate the candidate pool directly from the 1.5B SFT model, which we leave for future work.

\paragraph{Limitations.}
The price of this simplicity is that many design choices are heuristic.
The six evaluation dimensions, the aggregation into wrongness and
absurdity, and the particular form of $w_{\text{raw}, i}$ are motivated by
error analysis and fixtures, not by a formal theory of optimal
verification or sample selection.
We cannot claim that this weighting scheme is mathematically justified or
provably better than simpler baselines; our evidence is empirical and
limited in scope.
Moreover, the current pipeline is purely \emph{offline}: the verifier is
applied once to a fixed pool of completions, and DPO is run on the
resulting preference pairs.
We do not yet close the loop with online RL or allow the verifier to
adapt as the policy improves.
Finally, experiments are restricted to a 1.5B-parameter model and a small
set of benchmarks, so the scalability of the approach remains to be
validated.

\paragraph{Future directions.}
Given these limitations, we see three natural extensions.
First, to put the weighting scheme on firmer ground, one could reinterpret
hard-negative mining as active learning or importance sampling and design
verifier scores that approximate quantities like expected policy
improvement, rather than entering the loss as a linear combination.
Second, instead of treating the MathVerifier as a fixed module, it could
be co-trained with the policy (or distilled into a compact PRM) and used
inside an online RL loop, turning our offline DPO stage into a genuinely
closed feedback system.
Third, the pattern of decomposed verification plus hard-negative mining
plus weighted preference learning appears generic and may transfer to
other structured reasoning domains (e.g., code or formal proofs), where
partial programmatic checks are available.
In that sense, the main contribution of this work is the \emph{pipeline
architecture} itself, more than any single accuracy number.

\section{Conclusion}
\label{sec:conclusion}

This paper set out from a simple question:
\emph{can we move beyond treating math solutions from LLMs as merely
``correct'' or ``incorrect,'' and instead use the structure of their
reasoning, especially hard negatives, to improve them?}
To answer this, we designed and implemented a three-stage pipeline:
(i) supervised fine-tuning on MetaMathQA-style chain-of-thought data,
(ii) a lightweight MathVerifier that scores reasoning along six
interpretable dimensions and exposes aggregate wrongness/absurdity
scores, and (iii) a verifier-guided, weighted DPO stage that focuses
preference learning on high-value hard negatives.

Table~\ref{tab:gsm8k-main} summarizes the main quantitative results.
Starting from the \texttt{Qwen2.5-1.5B-Instruct} base model, SFT on
MetaMathQA provides a strong math-reasoning baseline; adding
verifier-guided DPO yields consistent (though modest) gains on standard
benchmarks such as GSM8K and MATH, in line with what is typically observed
for DPO-style post-training.

For a reader who only needs the key takeaways, the contributions of this
work can be summarized as follows:
\begin{itemize}
  \item \textbf{A decomposed math verifier.}
        We propose a dual-channel MathVerifier that evaluates reasoning
        along six dimensions and produces interpretable wrongness and
        absurdity scores, enabling fine-grained analysis of model
        failures beyond final-answer correctness.
  \item \textbf{Hard-negative-centered training.}
        We demonstrate how to use the verifier to mine high-value hard
        negatives and define a simple per-sample importance weight
        combining verifier error, model confidence, and perplexity, which
        can be plugged into a weighted DPO objective.
  \item \textbf{A practical, compute-conscious pipeline.}
        The entire framework operates in an offline SFT+DPO setting,
        avoids large PRMs and LLM-as-a-judge loops, and runs on a
        1.5B-parameter base model, making it accessible under realistic
        compute and engineering budgets.
\end{itemize}

What is missing so far are larger-scale experiments and a more principled
understanding of the weighting and verification schemes.
In particular, we have not yet: (i) validated the approach on larger
models and more diverse benchmarks, (ii) closed the loop with an online
RL stage, or (iii) connected the importance weights to any formal
active-learning or variance-reduction criterion.
These are natural directions for future work.

Overall, the main lesson from this work is that \emph{how} we organize
the training pipeline matters as much as the raw model size or dataset:
even a relatively simple verifier and heuristic weighting scheme can
already make DPO more targeted and interpretable, especially when one
intentionally centers hard negatives rather than treating them as a
byproduct.
We hope this perspective will be useful to practitioners who want to
improve mathematical reasoning without committing to the full cost of
end-to-end RLHF with large reward models.

\paragraph{Reproducibility.}
Code for the full pipeline (SFT, MathVerifier, and weighted DPO training) will be released at \url{https://github.com/HaochengLu/HardNeg-DPO} upon publication.

\section*{Ethics Statement}
This work focuses on improving mathematical reasoning in language models using publicly available datasets and open-weight base models.
We do not collect or use any private or personally identifiable data.
The training pipeline does not involve human subjects or human annotation.
We acknowledge that improvements in mathematical reasoning could in principle be misused (e.g., to generate plausible but subtly wrong solutions for academic dishonesty); however, the same verifier technology that enables our training pipeline can also be deployed as a detection tool for such misuse.

\bibliographystyle{IEEEtran}
\bibliography{references}

\end{document}